\documentclass[runningheads]{llncs}
\usepackage[T1]{fontenc}
\usepackage{amsmath}
\usepackage{graphicx}
\usepackage{svg}
\usepackage{tabularx,booktabs}
\newcommand{\repeatthanks}{\textsuperscript{\thefootnote}}
\usepackage[
colorlinks=true,
urlcolor=blue,
linkcolor=green
]{hyperref}
\newcolumntype{Y}{>{\centering\arraybackslash}X}
\begin{document}
\title{Neural Cellular Automata for Lightweight, Robust and Explainable Classification of White Blood Cell Images}
\titlerunning{Neural Cellular Automata for Image Classification}
% If the paper title is too long for the running head, you can set
% an abbreviated paper title here
%
\author{Michael Deutges \inst{1,2}\thanks{equal contribution} 
\and Ario Sadafi \inst{1,3}\repeatthanks 
\and Nassir Navab \inst{3,4} \and Carsten Marr \inst{1,5}}
% index{Deutges, Michael} 
% index{Sadafi, Ario} 
% index{Marr, Carsten} 
% index{Navab, Nassir} 
\authorrunning{M. Deutges and A. Sadafi et al.}
% First names are abbreviated in the running head.
% If there are more than two authors, 'et al.' is used.
%
\institute{Institute of AI for Health, Helmholtz Zentrum München – German Research Center for Environmental Health, Neuherberg, Germany \and
Faculty of Mathematics, Technical University Munich, Munich, Germany \and
Computer Aided Medical Procedures, Technical University of Munich, Munich, Germany \and Computer Aided Medical Procedures, Johns Hopkins University, USA \and Helmholtz AI, Helmholtz Zentrum München - German Research Center for Environmental Health, Neuherberg, Germany}

\maketitle              % typeset the header of the contribution
\begin{abstract}
Diagnosis of hematological malignancies depends on accurate identification of white blood cells in peripheral blood smears. Deep learning techniques are emerging as a viable solution to scale and optimize this process by automatic cell classification. However, these techniques face several challenges such as limited generalizability, sensitivity to domain shifts, and lack of explainability. Here, we introduce a novel approach for white blood cell classification based on neural cellular automata (NCA). We test our approach on three datasets of white blood cell images and show that we achieve competitive performance compared to conventional methods. Our NCA-based method is significantly smaller in terms of parameters and exhibits robustness to domain shifts. Furthermore, the architecture is inherently explainable, providing insights into the decision process for each classification, which helps to understand and validate model predictions. Our results demonstrate that NCA can be used for image classification, and that they address key challenges of conventional methods, indicating a high potential for applicability in clinical practice.

\end{abstract}
\section{Introduction}
The diagnosis of hematological disorders heavily relies on the microscopic examination of blood cells in laboratory settings. Among the challenges that cytologists encounter is the identification of relevant white blood cells under the microscope for diagnosing leukemia, which ranks among the most lethal hematological disorders. Certain types of leukemia, such as acute promyelocytic leukemia (APL), pose significant challenges as they demand urgent attention due to the potential for life-threatening bleeding or coagulation \cite{gill2021characteristics}.

Identifying relevant cells, known as blast cells, under the microscope is an essential initial step in diagnosing leukemia subtypes, including APL. This task is time consuming and burdensome. 
Recent advancements in deep learning methods have introduced automatic tools for cytologists, enabling them to expedite the process of locating and classifying these critical cells. As a result, the diagnosis can become faster and more robust, streamlining the workflow in leukemia laboratories.
Matek et al. \cite{matek2021highly,matek2019human} suggested a highly accurate classifier for bone marrow cells and cells from peripheral blood, demonstrating performance comparable with trained cytologists.
Eckardt et al. proposed deep learning methods for diagnosis of acute myeloid leukemia (AML) subtypes \cite{eckardt2022deep,eckardt2022deep2}. In a similar work, Sidhom et al. focused on diagnosis of APL from other subtypes of AML \cite{sidhom2021deep}. 

These methods are typically trained and validated on datasets gathered from a single source, rendering them susceptible to domain shifts \cite{guan2021domain}. To enhance cross-domain performance, various domain adaptation methods have been proposed. These include extracting domain-invariant features \cite{salehi2022unsupervised} or employing continual learning techniques \cite{sadafi2023continual}, which facilitate regular model updates to address challenges introduced by new domains. 

Explaining the decisions of convolutional neural networks (CNNs) presents a significant challenge in the applicability of deep learning methods and is crucial for computer aided diagnosis systems \cite{hehr2023explainable,sadafi2023pixel}. This explanation is typically achieved through pixel or feature attributions, where certain parts of the image are identified as important by the explanation method \cite{sadafi2023pixel}. The interpretation of these clues relies on the observer, who examines the explanations to form an understanding of the system's inner workings.

Neural cellular automata (NCA) are gradually emerging as a lightweight, robust and input-invariant solution for image generation or segmentation tasks. Growing NCA are able to generate, maintain and regenerate complex shapes \cite{mordvintsev2022growing}. In the medical domain, Kalkhof et al. \cite{kalkhof2023med} have proposed Med-NCA for segmentation of different organs in T1-weighted hippocampus and T2-MRI datasets. It demonstrates comparable performance with nnUNet \cite{isensee2018nnu}, while having a fraction of its parameters.

In this paper, we propose a method for explainable and robust single white blood cell classification utilizing a NCA backbone. Features from the single-cell images are extracted by the NCA and used for classification with a multi-layer perceptron. Our approach inherently offers explainability and demonstrates robustness against domain shifts when evaluated on three datasets collected from different medical centers, each with its specific laboratory procedures and staining techniques.% To foster reproducible research we are providing our source code at [left out for anonymity].

To the best of our knowledge, this is the first application of NCA for image classification. Our contribution is thus the methodological advancement of a new high-potential method and its application to a challenging biomedical classification task. To foster reproducible research we are providing our source code at \url{https://github.com/marrlab/WBC-NCA}.
\section{Methods}
Conventional deep learning models for image classification  involve two steps: feature extraction and classification. In our approach, we extract the features with the use of NCA. Unlike traditional methods that rely on a series of convolutions, activation functions, and pooling operations, NCA apply a local update rule iteratively over a fixed number of steps, allowing each cell to aggregate information from a wider context. This local architecture of NCA, which update cells based on their immediate surroundings, ensures lightweight storage and fast inference without compromising performance.

\subsection{Neural Cellular Automata Architecture}
A NCA can be defined as a tuple 
\begin{equation}
    \textrm{NCA}=\langle S,f \rangle
\end{equation}
where \(S \in \mathbf{R}^{64 \times 64 \times n}\) denotes the seed and \(f:\mathbf{R}^{3\times 3 \times n}\rightarrow \mathbf{R}^n\) is the transition function, which is applied iteratively starting from \(S\). The dimensions of the seed imply the number of individual cells  \(c \in \mathbf{R}^n\) which, arranged in a \(64\) by \(64\) grid, comprise the domain our NCA operates on. We refer to each \(64\times 64\) slice of the domain as a channel.
In our case, \(S\) consists of the RGB image \(\in \mathbf{R}^{64\times 64 \times 3}\) and \(n-3\) channels of zeros.

The transition function updates each cell \(c\) based on its \(3 \times 3\) neighborhood denoted by \(N_c\) and can be divided into two parts: perception \(f_p\) and update \(f_u\).
\begin{align}\label{eq:2}
    &f:\mathbf{R}^{3\times 3 \times n}\rightarrow \mathbf{R}^n, &&N_c \mapsto f_u(f_p(N_c))\\
    &f_p:\mathbf{R}^{3\times 3 \times n}\rightarrow \mathbf{R}^{3n}, &&N_c \mapsto 
    \left(c,N_c*k_1,N_c*k_2\right)^T\\
    &f_u:\mathbf{R}^{3n}\rightarrow \mathbf{R}^{n}, &&f_p(N_c) \mapsto W_2\max(W_1f_p(N_c)+b_1,0)+b_2
\end{align}
We call \(f_p(N_c)\) the perception vector. It consists of two channel-wise convolutions with kernels \(k_1\) and \(k_2\) and the identity of the cell \(c\). The update function \(f_u\) is comprised of two fully connected layers, parameterized by \(W_1\), \(W_2\), \(b_1\), and \(b_2\), with a Rectified Linear Unit (ReLU) activation in between. The convolution kernels \(k_1\) and \(k_2\) and the weights \(W_1\), \(W_2\) and biases \(b_1\), \(b_2\) of the linear layers are trainable parameters.

\begin{figure}[t]
\centering
\includegraphics[width=\textwidth,page=1,trim=1cm 2cm 3cm 1.5cm,clip]{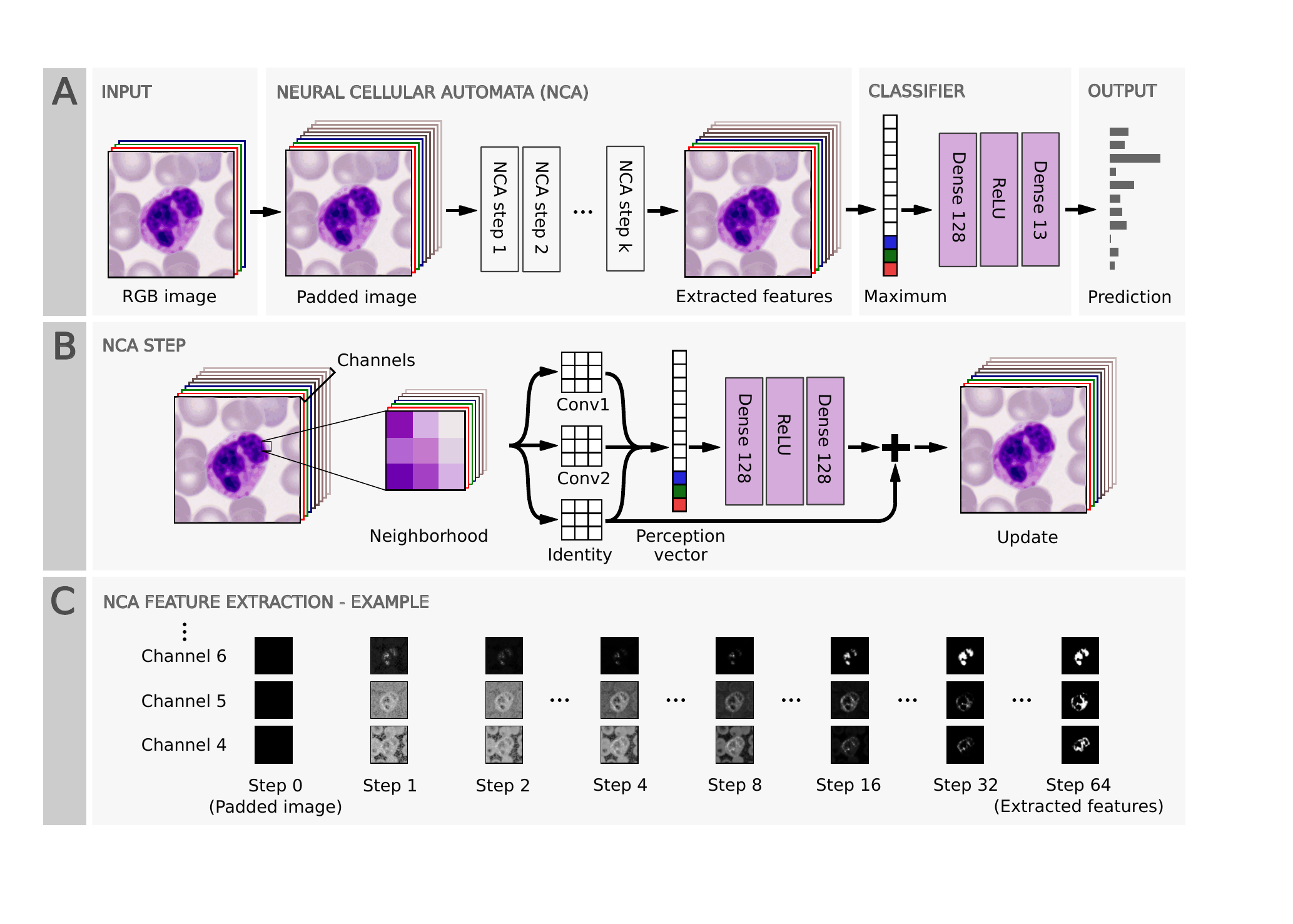}
\caption{Neural cellular automata (NCA) can be used for the accurate classification of single white blood cells in patient blood smears. \textbf{A:} Our approach consists of four steps: i) image padding to increase the number of channels, ii) \(k\) NCA update steps to extract features from the image that manifest in the hidden channels, iii) pooling via channel-wise maximum, and iv) a fully connected network to classify the image. \textbf{B:} The NCA step updates each cell based on its immediate surroundings according to equations \ref{eq:2} - \ref{eq:5}. \textbf{C:} Training the model end-to-end allows the NCA to learn an update rule that extracts useful features.} 
\label{fig:fig1}
\end{figure}

The transition function defines the cell update at time step \(t\) according to
\begin{equation}\label{eq:5}
    c^{t+1}=c^t+\delta f(N_{c^t})
\end{equation}
where \(\delta\) is randomly set to \(0\) or \(1\) for each cell in each step, i.e. only approximately \(50\%\) of cells get updated in each step. This stochastic activation acts as regularization, enhancing the model's robustness and improving generalization. The architecture and NCA step including perception and cell update is illustrated in Figure \ref{fig:fig1}.

\subsection{NCA for Image Classification}
To use NCA for a classification task we perform \(k\) update steps, which results in extracted features that form in the additional channels (Figure \ref{fig:fig1}C). Each channel is condensed via taking the maximum, which yields a rich feature vector denoted by \(v\). We arrive at the prediction by feeding this vector into a two layer fully connected network denoted as the classifier \(g\).
\begin{equation}
    g:\mathbf{R}^{n}\rightarrow [0,1]^{13},\hspace{0.5cm}g(v) \mapsto \sigma(W_4\max(W_3v+b_3,0)+b_4)
\end{equation}
with \(\sigma(z)=\frac{1}{1+e^{-z}}\) and weights and biases \(W_3,W_4,b_3,b_4\). The output \(g(v)\) corresponds to the predictions of the model for the 13 different cell types in our datasets. 

In summary, our model is structured into four steps (Figure \ref{fig:fig1}):
\begin{enumerate}
    \item \textbf{Image Padding:} We pad the image to the desired number of channels. Additional channels serve as memory, which enables the NCA to learn more complex patterns.
    \item \textbf{NCA Update Steps:} We perform \(k\) NCA update steps to extract features which manifest in the hidden channels.
    \item \textbf{Feature Aggregation:} Channel-wise maximum is taken to condense the extracted features into a vector.
    \item \textbf{Classification:} This condensed vector is fed into a two layer fully connected neural network which yields the class predictions.
\end{enumerate}
During training, we apply the cross-entropy loss function to the class prediction and the corresponding ground truth, backpropagating through both the classifier network and the NCA over time. This end-to-end training scheme enables the NCA to learn an update rule that, when applied for \(k\) steps, extracts features from the image which are effective for classification.

\subsection{Explainability via Layer-wise Relevance Propagation}
To gain insights into the model's decision making, we use layer-wise relevance propagation \cite{Montavon2019} on the fully connected layers of the classifier network to attribute relevance to each feature extracted by the NCA. We apply the epsilon rule described in \cite{Montavon2019}. Relevance values are propagated through each layer according to
\begin{equation}
    R^{L-1}_j=\sum_k \frac{a_jw_{jk}}{\epsilon + \sum_ja_jw_{jk}}R_k^{L},
    \label{eq:6}
\end{equation}
where \(\epsilon\) is a parameter to control sparsity of the explanations, \(a_j\) are the activations of the respective neurons and \(w_{jk}\) corresponds to the weights in the layer. If \(\epsilon = 0\), this definition fulfills \(\sum_jR^{L}_j=p(x)\) for every layer \(L\), where \(p(x)\) is the prediction of the model. Relevance values \(R_j\) correspond to the contribution of feature \(j\) to the prediction of the specified class.

\section{Experiments and Results}
\subsection{Datasets}
We use three different datasets to evaluate our method:
\begin{itemize}
    \item \textbf{Matek-19} is a collection of over 18,000 annotated white blood cells from 200 individuals where half of the subjects are affected by AML. The data has been collected at the Munich university hospital and is publicly available \cite{matek2019human}. It consists of 15 classes, and the images have a size of $400 \times 400$ pixels, which corresponds to $29\times29$ micrometers.
    \item \textbf{INT-20} is an in-house dataset and consists of around 42,000 images from 18 different classes with a resolution of $288\times288$ pixels or $25\times25$ micrometers.
    \item \textbf{Acevedo-20} is a publicly available dataset of around 17,000 single-cell images from healthy individuals collected at the Hospital Clinic of Barcelona \cite{acevedo2020dataset}. The images are categorized in 8 different classes and have a resolution of $360\times363$ pixels or $36\times36.3$ micrometers.
\end{itemize}
\begin{figure}
\centering
\includegraphics[width=\textwidth,page=3,trim=0cm 2cm 2cm 2.5cm,clip]{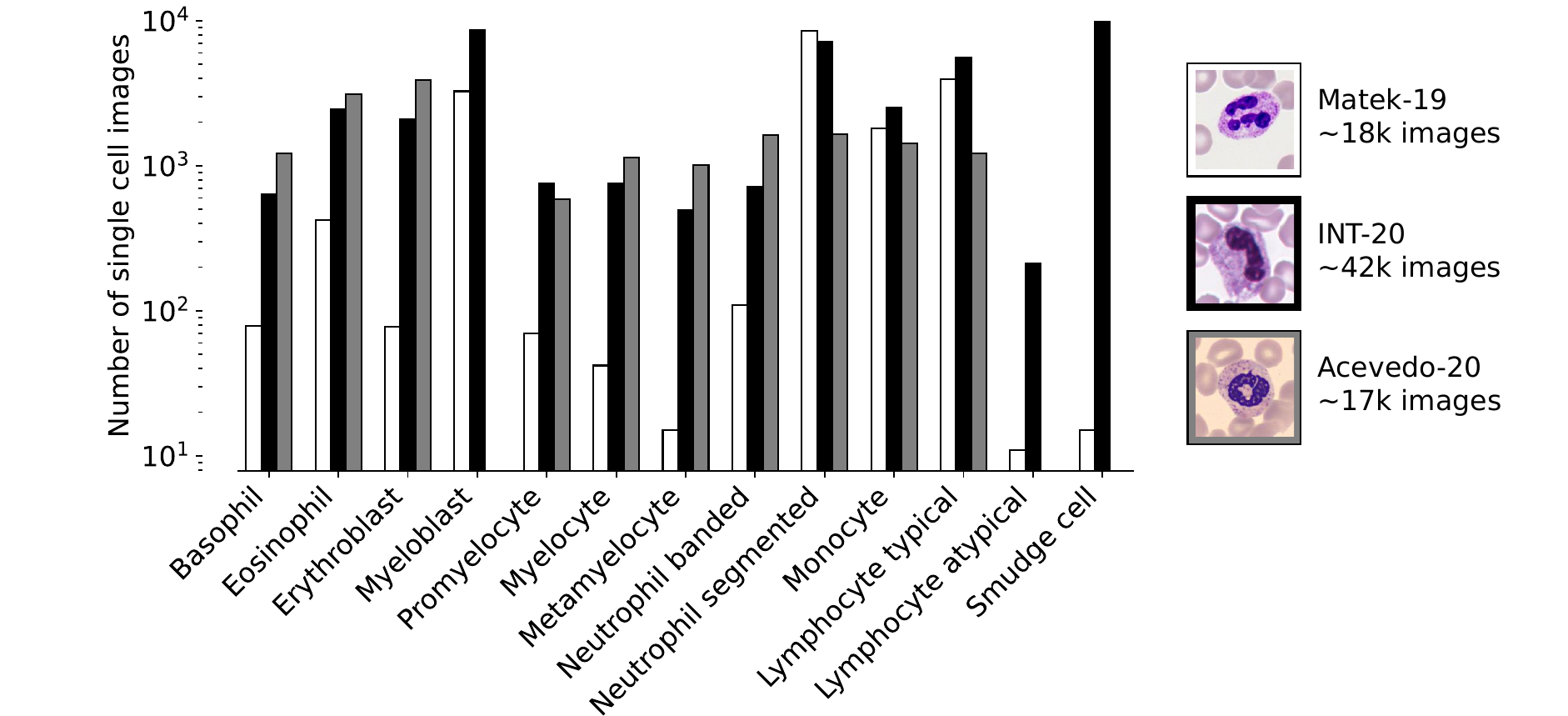}
\caption{Class distribution for the three datasets.} 
\label{fig:fig3}
\end{figure}
The class definitions of these three datasets are harmonized by a medical expert resulting in 13 commonly defined classes. Figure \ref{fig:fig3} shows the distribution of cell types for the three datasets. 

We resampled the images to a resolution of \(64 \times 64\) as a high resolution would significantly increase time and RAM requirements for training the NCA.
\subsection{Implementation details}
\subsubsection{Model Choice}
The standard configuration of our NCA model has \(n=128\) channels. A large number of channels provides cells with a bigger memory capacity and allows the NCA to learn a more advanced update rule.
\begin{figure}
\centering
\includegraphics[width=0.6\textwidth,page=2,trim=0cm 0cm 0cm 0cm,clip]{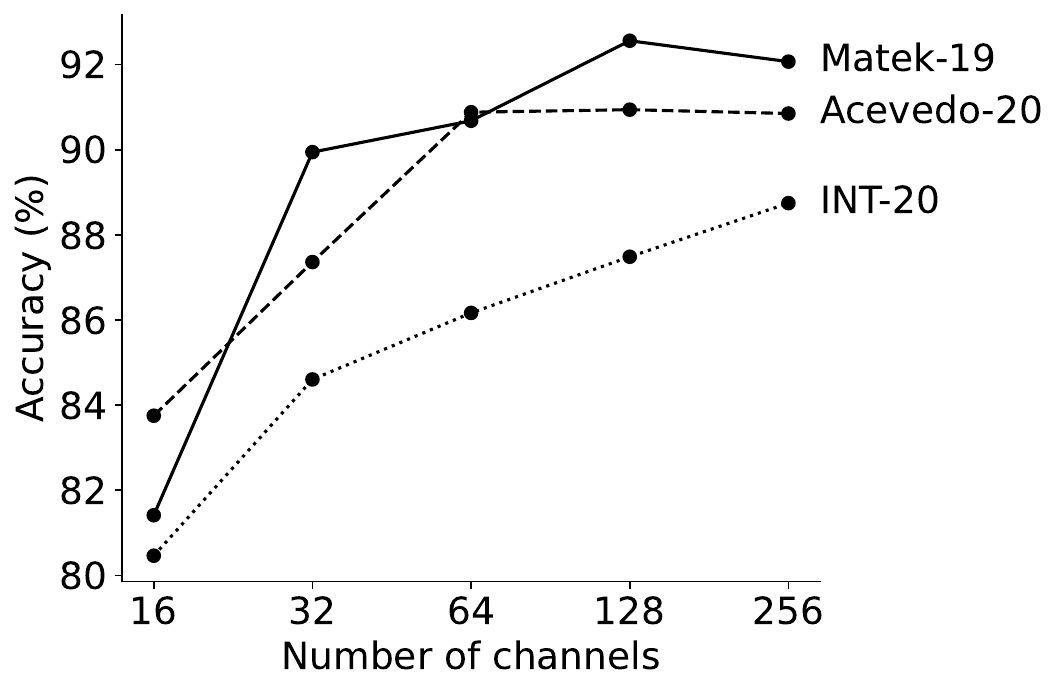}
\caption{NCA classification accuracy saturates at 128 channels for two out of three white blood cell datasets.} 
\label{fig:fig2}
\end{figure}
We tested the classification accuracy on the three datasets for models with different numbers of channels to validate this choice (Figure \ref{fig:fig2}). 
The fully connected layers in the transition function have a hidden size of 128, which is consistent with other NCA models \cite{kalkhof2023med,mordvintsev2022growing}. A total of \(k=64\) steps are performed before the features are aggregated and propagated through the classifier network. We found that the number of steps seems to correlate positively with the accuracy. We opted for 64 steps as a compromise between performance and training time and RAM requirements, respectively. 

\subsubsection{Training}
We used an Adam optimizer with a learning rate of 0.0004 and a \(\beta_1\) of 0.9 and \(\beta_2\) of 0.999 together with an exponential learning rate decay with weight 0.9999.
To treat the problems arising from an imbalanced dataset, we used random over- and undersampling together with random rotation and flip augmentations of the training images to have a uniform distribution of classes in the training process. Each model was trained with a batch size of 16 for 32 epochs, after which the validation loss did not change significantly anymore. In general, we observed our model to be notably less prone to overfitting compared to conventional methods. This might be a consequence of the small number of parameters which sums up to just about 86k.

\subsection{Results}
We compare the performance in terms of accuracy on the three datasets against two baselines. Matek et. al \cite{matek2019human} proposed a method based on a ResNeXt \cite{xie2017aggregated} architecture for the classification of the Matek-19 dataset. The model binaries are available, which we used for testing on the three datasets. In order to compare its performance when trained on the other datasets, we are employing the same ResNeXt architecture to train and test. 
\begin{table}[h]
\caption{When trained on data from one domain, and tested on data from a different domain, our NCA model outperforms other baselines, while only having a fraction of the parameters. The accuracy of our NCA method is reported against the baselines. Mean and standard deviation are computed from five independent runs.% The results of the ResNeXt architecture trained on Matek-19 use the publicly available model binaries trained by Matek et al.  
}\label{tab1}
\begin{tabularx}{\textwidth}{Y|Y||Y|Y|Y}
& & NCA & \multicolumn{1}{c|}{ResNeXt}& AE-CFE \\ 
Trained on & Tested on & $\sim$86k pa. & \multicolumn{1}{c|}{$\sim$25M pa.} & $\sim$3.9M pa. \\ \hline
 & Matek-19 & 92.6$\pm$0.6 & \textbf{96.1} & 83.7$\pm$0.5 \\ \cline{2-5} 
Matek-19 & Acevedo-20 & \textbf{43.9$\pm$1.6} & 8.1 & 21.9$\pm$0.4 \\ \cline{2-5} 
 & INT-20 & 24.3$\pm$5.1 & 29.5 & \textbf{48.4$\pm$0.2} \\ \hline
 & Matek-19 & 32.2$\pm$9.1 & 7.3$\pm$3.1 & \textbf{45.1$\pm$0.5} \\ \cline{2-5} 
Acevedo-20 & Acevedo-20 & \textbf{90.6$\pm$0.2} & 85.7$\pm$2.4 & 65.2$\pm$0.5 \\ \cline{2-5} 
 & INT-20 & 13.6$\pm$2.8 & 8.1$\pm$1.4 & \textbf{21.0$\pm$0.5} \\ \hline
 & Matek-19 & 50.0$\pm$11.1 & 49.0$\pm$6.3 & \textbf{73.2$\pm$0.1} \\ \cline{2-5} 
INT-20 & Acevedo-20 & \textbf{46.9$\pm$4.7} & 16.9$\pm$1.6 & 31.8$\pm$0.4 \\ \cline{2-5} 
 & INT-20 & 88.0$\pm$0.3 & \textbf{88.7$\pm$1.5} &  65.6$\pm$0.5
\end{tabularx}
\end{table}
Additionally, we are comparing against AE-CFE proposed by Salehi et al. \cite{salehi2022unsupervised}, which uses an autoencoder trained on all three datasets to extract domain invariant features, which are later used in a random forest to classify single white blood cell images.

Our model outperforms the ResNext architecture in 6 out of 9 experiments. Furthermore, our model outperforms AE-CFE in all experiments where the training and testing set come from the same center. The advantage of AE-CFE in the other experiments is a result of the pretraining on all three domains. 

\subsection{Explainability}
\begin{figure}
\centering
\includegraphics[width=\textwidth,page=4,trim=2cm 5cm 5cm 5cm,clip]{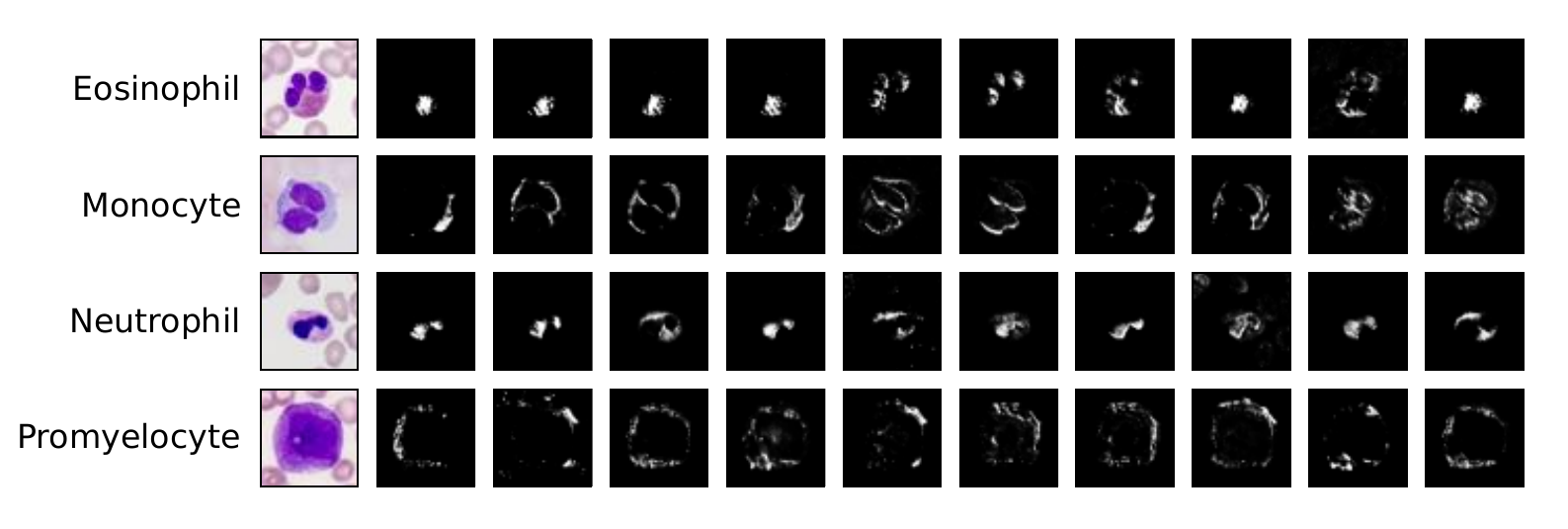}
\caption{Example of ten features (columns) extracted by the NCA for an exemplary eosinophil, monocyte, neutrophil, and promyelocyte. The highest relevance scores have been attributed by layer-wise relevance propagation on the fully connected layers of the classifier network. The channel activations are located within the regions of the white blood cell indicating the model's ability to correctly identify and analyze pertinent aspects of the input data.} 
\label{fig:fig4}
\end{figure}

Our NCA model offers explainability by design. Features are extracted in the channels and offer insights into the decision process. A visualization of the channels allows for easy identification of biases, e.g. if the model were to make decisions based on activations that are located not within the region of the cell but in the background. An example is shown in Figure \ref{fig:fig4}. Additionally, the unique architecture of the NCA yields features which are more interpretable by humans. The extraction via local cell updates aggregated over time results in features which tend to be more connected regions of the image. In our scenario, these regions could include the entire cell, its nuclei, cytoplasm parts, or more subtle sub-cellular structures.
\section{Conclusion}
In this study, we demonstrated the potential of neural cellular automata as a tool for image classification, addressing key challenges faced by conventional methods in clinical practice. By providing a combination of performance, robustness, and explainability, our approach provides an alternative to improved diagnostic tools in the field of hematological diseases \cite{eckardt2022deep,eckardt2022deep2,hehr2023explainable,matek2021highly,matek2019human,sidhom2021deep}.

The inherent explainability of NCA simplifies the process of interpreting the inner workings for observers, offering a distinct advantage compared to the noisy pixel attributions typically used for explaining CNNs \cite{sadafi2023pixel}.

Lastly, the lightweight architecture allows for application in less developed and remote areas for various diagnostic tasks without requiring access to high-end hardware to run large models.

%\begin{credits}
\subsubsection{Acknowledgements.} We would like to thank John Kalkhof for the inspiring discussions. C.M. acknowledges funding from the European Research Council (ERC) under the European Union's Horizon 2020 research and innovation program (Grant Agreement No. 866411 \& 101113551) and support from the Hightech Agenda Bayern.
\subsubsection{Disclosure of Interests.} The authors have no competing interests to declare that are relevant to the content of this article.
%\end{credits}

\bibliographystyle{splncs04}
\bibliography{Paper-3422}
\nocite{tesfaldet2022attention,florindo2021cellular,randazzo2020self,yecsil2023novel}
\end{document}